\documentclass{article}

\usepackage{fancyheadings}
\usepackage{enumerate}
\usepackage{mdwlist}
\usepackage{natbib}
\usepackage{url}

\begin{document}

\pagestyle{fancy}
\cfoot{\scriptsize{This is the TempEval-3 proposal, not the main paper, which is in the SemEval-2013 proceedings.}}

\title{\textsc{TempEval-3}: Evaluating Events, Time Expressions, and Temporal Relations}
\author{Naushad UzZaman$^{\ast}$, Hector Llorens$^{\dagger}$, James Allen$^{\ast}$, \\ Leon Derczynski$^{\ddagger}$, Marc Verhagen$^{\diamond}$ and James Pustejovsky$^{\diamond}$ \\ \\ 
\small $\ast$: \texttt{\{naushad,james\}@cs.rochester.edu} University of Rochester, NY, USA\\
\small $\dagger$: \texttt{hllorens@dlsi.ua.es} University of Alicante, Spain\\
\small $\ddagger$: \texttt{leon@dcs.shef.ac.uk} University of Sheffield, UK\\
\small $\diamond$: \texttt{\{jamesp,marc\}@cs.brandeis.edu} Brandeis University, MA, USA\\
}
\date{}

\maketitle

\begin{abstract}
In this proposal, we describe the forthcoming TempEval-3 task which is currently in preparation for the SemEval-2013 evaluation exercise. The aim of TempEval is to advance research on temporal information processing. TempEval-3 follows on from previous TempEval events, incorporating: a three-part task structure covering event, temporal expression and temporal relation extraction; a larger dataset; and single overall task quality scores.
\end{abstract}

\section{Introduction}
The TempEval task was added as a new task in SemEval-2007~\citep{TempEval1}, focusing on the identification of temporal relations. The automatic identification of all temporal referring expressions, events, and temporal relations within a text is the ultimate aim of research in this area. The area is too broad to address completely in a first evaluation challenge, and a staged approach was taken instead. TempEval (henceforth TempEval-1) was an initial evaluation exercise based on three fixed-scope tasks (identifying links between: events and timexes in the same sentence; events and document creation time DCT; main events in successive sentences) that were considered realistic both from the perspective of assembling resources for development and testing and from the perspective of developing systems capable of addressing the tasks.

TempEval-2~\citep{TempEval2} extended TempEval-1, growing into a multilingual task, and consisting of six subtasks rather than three. This included event and timex extraction, as well as the three relation tasks from TempEval-1, with the addition of a relation task where one event subordinates another.

Temporal annotation is a time-consuming task for humans, which has limited the size of annotated data in previous TempEval exercises. Current systems, however, are performing close to the inter-annotator reliability for entity recognition. This suggests that larger corpora could be built from automatically annotated data with minor human reviews. As part of TempEval-3, we explore whether there is value in adding a large automatically created silver standard to a hand-crafted gold standard. 

Automatic performance on temporal relation annotation is still limited; in TempEval-2, systems achieved an above-baseline error reduction of less than 20\% for most tasks. This suggests that the temporal relation problem is still open and remains a topic of intensive contemporary research.

With these points in mind, this paper describes the next upcoming temporal evaluation shared task -- TempEval-3 -- to be held with SemEval-2013. 

\section{TempEval changes}
As proposed, TempEval-3 is a follow-up to TempEval-1 and 2. TempEval-3 differs from its ancestors in the following respects: 

\begin{enumerate}[(i)]
\item size of the corpus: the dataset used comprises about 500K tokens of silver standard data and about 100K tokens of gold standard data for training, compared to the corpus of roughly 50K tokens corpus used in TempEval 1 and 2;
\item temporal relation task: the temporal relation classification tasks are to be performed from raw text, i.e. participants need to extract events and temporal expressions first, determine which ones to link and then obtain the relation types;
\item tasks not independent: participants must annotate temporal expressions and events in order to do the relation task;
\item temporal relation types: the full set of temporal interval relations in TimeML~\citep{pustejovsky2005specification} is used, rather than the reduced set used in earlier TempEvals;
\item annotation: most of the corpus was automatically annotated by the state-of-the-art systems from TempEval-2, a portion of the corpus, including the test dataset, that is human reviewed;
\item evaluation: we will report a temporal awareness score for evaluating temporal relations, to help to rank systems with a single score.
\end{enumerate}

\pagebreak 

\section{Tasks}

The tasks proposed for TempEval-3 are: 

\subsection{Task A: Temporal expression extraction and normalization}
Determine the extent of the time expressions in a text as defined by the TimeML TIMEX3 tag. In addition, determine the value of the features TYPE and VAL. The possible values of TYPE are time, date, duration, and set; the value of VAL is a normalized value as defined by the TIMEX3 standard. The main attribute to annotate is VAL.

\subsection{Task B: Event extraction}
As in TempEval-2, participants will determine the extent of the events in a text as defined by the TimeML EVENT tag. In addition, systems may determine the value of the features CLASS, TENSE, ASPECT, POLARITY, MODALITY and also identify if the event is a main event or not. The main attribute to annotate is CLASS.

\subsection{Task C: Annotating temporal relations}
Identify the pairs of temporal entities (events or temporal expressions) that have a temporal link and classify the temporal relation between them as a TLINK.  Possible pairs of entities that can have a temporal link are: 
(i) event and temporal expressions in the same sentence, (ii) event and document creation time, (iii) main events of consecutive sentences and (iv) pairs of events in the same sentence. For this task, we now require that the participating systems determine which entities need to be linked. 

The relation labels will be same as in TimeML, i.e.: \textsc{before, after, includes, is-included, during, simultaneous, immediately after, immediately before, identity, begins, ends, begun-by} and \textsc{ended-by}.

\subsection{Task selection}
Participants may choose to do task A, B, or C. Choosing task C (relation annotation) entails doing tasks A and B (interval annotation). However, a participant may perform only task C by applying existing tools to carry out tasks A and B.

\pagebreak

\section{Dataset creation}

In TempEval-3, we release new data, as well as significantly reviewing and modifying existing corpora.

\subsection{Reviewing Existing Corpora} \label{reviewedexisting}
We considered the existing TimeBank~\citep{TimeBank}, TempEval-1, TempEval-2 and AQUAINT\footnote{\url{http://timeml.org/site/timebank/timebank.html}} data for review in TempEval-3. TimeBank v1.2, TempEval-1 and TempEval-2 had the same documents but different relation types and sometimes different sets of events. We will refer to this body of temporally-annotated newswire documents as TimeBank. 

For both TimeBank and AQUAINT, we cleaned up the formatting for all files making it easy to review and read, made all files XML and TimeML schema compatible and added some missing events and temporal expressions. In AQUAINT, we added the temporal relations between event and DCT (document creation time), which was missing for many documents in that corpus. In particular, for the TimeBank documents, we borrowed the events from the TempEval-2 corpus and the temporal relations from the TimeBank corpus, which contains a full set of temporal relations (TempEval-2 used a simpler, coarse-grained set of temporal relations). 

A standard datafile format has been adopted, which is a subset of valid ISO-TimeML. It begins with an outer \texttt{TimeML} element as normal. The document name is contained in a child \texttt{DOCID} element, any newswire preamble in an\\ optional \texttt{EXTRAINFO} element, headline in an optional \texttt{TITLE} element, the document timestamp in a \texttt{DCT} element (usually with an ID of \texttt{t0}) and the main body of the text to be annotated in a \texttt{TEXT} element. For example:

\vspace{2mm}

\footnotesize

\texttt{<?xml version="1.0" ?>}

\texttt{<TimeML xmlns:xsi="http://www.w3.org/2001/XMLSchema-instance"\\ xsi:noNamespaceSchemaLocation="http://timeml.org/timeMLdocs/TimeML\_1.2.1.xsd">}

\texttt{<DOCID>XIN\_ENG\_20061119.0021</DOCID>}

\texttt{<DCT>HANOI, <TIMEX3 functionInDocument="CREATION\_TIME" temporalFunction="false" tid="t0" type="TIME" value="2006-11-19">Nov. 19 , 2006</TIMEX3> (Xinhua)</DCT>}

\texttt{<TITLE>URGENT: Russia, US sign agreement on WTO deal in Vietnam</TITLE>}

\texttt{<TEXT>}

\texttt{Russia and the United States Sunday <EVENT aspect="NONE" class="OCCURRENCE"\\ eid="e1" eiid="ei1" polarity="POS" pos="VERB" tense="PAST">signed</EVENT> a\\ bilateral <EVENT aspect="NONE" class="OCCURRENCE" eid="e2" eiid="ei2"\\ polarity="POS" pos="NOUN" tense="PAST">agreement</EVENT> on Russia's accession to\\ the World Trade Organization (WTO) on the sidelines of the ongoing Asia- Pacific\\ Economic Cooperaiton Economic Leaders' Meeting in Hanoi.}

\texttt{</TEXT>}

\texttt{<TLINK eventInstanceID="ei1" lid="l1" relType="NONE" relatedToTime="t0"/>}

\texttt{<TLINK eventInstanceID="ei2" lid="l2" relType="NONE"\\ relatedToEventInstance="ei1"/>}

\texttt{</TimeML>}

\normalsize

\subsection{Automatically Creating New Large Corpora}
A large portion of the TempEval-3 data is automatically generated, using a temporal merging system. We collected the half-million token text corpus from English Gigaword\footnote{\url{http://www.ldc.upenn.edu/Catalog/catalogEntry.jsp?catalogId=LDC2011T07}}. We automatically annotated this corpus using TIPSem, TIPSem-B~\citep{TIPSem} and TRIOS~\citep{TRIOS}. These systems were re-trained on the TimeBank and AQUAINT corpus, using the TimeML temporal relation set. We then merged these three state-of-the-art system outputs using our merging algorithm~\citep{uzzaman2012merging}. In our merging configuration, all entities and relations suggested by the best system (TIPSem) are added to the merge output. Suggestions from two other systems (TRIOS and TIPSem-B) are added to the merge output if they are supported by at least 2 of the 3 systems overall. The weights used in our configuration are: TIPSem 0.36, TIPSemB 0.32, TRIOS 0.32. 

This automatically created corpus is referred as \textbf{silver} data. A portion of the silver data is in the process of human reviewing for release as additional gold training data, in addition to reviewed and re-curated versions of TimeBank and AQUAINT. The parts described in Table~\ref{tempeval3-corpus} comprise our released dataset.

\begin{table}[htdp]
\caption{Available corpus released for TempEval-3.  (*: reviewing in progress)}
\label{tempeval3-corpus}
\begin{center}
\begin{tabular}{lrcc}
\hline
\textbf{Corpus} & \textbf{Number of tokens} & \textbf{Purpose} & \textbf{Standard} \\ 
\hline 
TimeBank & 61~418 & Training & Gold \\ 
AQUAINT & 33~973 & Training & Gold \\ 
TempEval-3 Silver & 666~309 & Training & Silver \\ 
TempEval-3 Gold & 20~000* & Training & Gold \\ 
TempEval-3 Evaluation &  20~000* & Evaluation & Gold \\ 
\hline
\end{tabular}
\end{center}
\end{table}%

The exploration of the benefits of both very large automatically temporally annotated corpora (silver data) and of smaller human annotated/reviewed temporal annotated corpora (gold data) with our TempEval-3 release is left to task participants and to future research. 

\section{Evaluation}
Evaluation on tasks A and B will be a standard F-score (incorporating Precision and Recall metrics) on extents and F-score/Kappa on attributes on the response extents that overlap with the key extents. Evaluation on task C will be incorporated from our proposed graph-based evaluation metric (see~\citet{uzzaman2011temporal} for details). This metric uses temporal closure to reward relation annotations that are equivalent but distinct and then finds precision and recall. Our \textbf{temporal awareness score} is a combined measure of a system's performance (i.e. it evaluates how a system extracts events, temporal expressions and also identifies all temporal relations). 

\section{Conclusion}
We have described the task, dataset and evaluation style for TempEval-3. The event will be part of SemEval-2013. Training will begin in autumn 2012, and the evaluation period ends January 2013. Further information can be found on the task website\footnote{\url{http://www.cs.york.ac.uk/semeval-2013/}} and via the TempEval group list\footnote{\url{http://groups.google.com/group/tempeval}}.

\bibliographystyle{te}
\bibliography{te3tech}

\begin{thebibliography}{8}
\newcommand{\enquote}[1]{``#1''}
\providecommand{\natexlab}[1]{#1}
\providecommand{\url}[1]{\texttt{#1}}
\providecommand{\urlprefix}{URL }
\providecommand{\bibAnnoteFile}[1]{%
  \IfFileExists{#1}{\begin{quotation}\noindent\textsc{Key:} #1\\
  \textsc{Annotation:}\ \input{#1}\end{quotation}}{}}
\providecommand{\bibAnnote}[2]{%
  \begin{quotation}\noindent\textsc{Key:} #1\\
  \textsc{Annotation:}\ #2\end{quotation}}

\bibitem[{Llorens et~al.(2010)Llorens, Saquete, and Navarro}]{TIPSem}
Llorens, H., E.~Saquete, and B.~Navarro (2010), \enquote{{TIPSem (English and
  Spanish): Evaluating CRFs and Semantic Roles in TempEval-2}.} In
  \emph{Proceedings of the 5th International Workshop on Semantic Evaluation},
  284--291, Association for Computational Linguistics.
\bibAnnoteFile{TIPSem}

\bibitem[{Pustejovsky et~al.(2003)Pustejovsky, Hanks, Sauri, See, Gaizauskas,
  Setzer, Radev, Sundheim, Day, Ferro et~al.}]{TimeBank}
Pustejovsky, J., P.~Hanks, R.~Sauri, A.~See, R.~Gaizauskas, A.~Setzer,
  D.~Radev, B.~Sundheim, D.~Day, L.~Ferro, et~al. (2003), \enquote{{The
  TimeBank corpus}.} In \emph{Corpus Linguistics}, volume 2003, 40.
\bibAnnoteFile{TimeBank}

\bibitem[{Pustejovsky et~al.(2005)Pustejovsky, Ingria, Sauri, Castano, Littman,
  Gaizauskas, Setzer, Katz, and Mani}]{pustejovsky2005specification}
Pustejovsky, J., B.~Ingria, R.~Sauri, J.~Castano, J.~Littman, R.~Gaizauskas,
  A.~Setzer, G.~Katz, and I.~Mani (2005), \enquote{{The specification language
  TimeML}.} \emph{The Language of Time: A reader}, 545--557.
\bibAnnoteFile{pustejovsky2005specification}

\bibitem[{UzZaman and Allen(2010)}]{TRIOS}
UzZaman, N. and J.F. Allen (2010), \enquote{{TRIPS and TRIOS system for
  TempEval-2: Extracting temporal information from text}.} In \emph{Proceedings
  of the 5th International Workshop on Semantic Evaluation}, 276--283,
  Association for Computational Linguistics.
\bibAnnoteFile{TRIOS}

\bibitem[{UzZaman and Allen(2011)}]{uzzaman2011temporal}
UzZaman, N. and J.F. Allen (2011), \enquote{{Temporal Evaluation}.} In
  \emph{Proceedings of The 49th Annual Meeting of the Association for
  Computational Linguistics: Human Language Technologies (Short Paper),
  Portland, Oregon, USA}.
\bibAnnoteFile{uzzaman2011temporal}

\bibitem[{UzZaman et~al.(2012)UzZaman, Llorens, and Allen}]{uzzaman2012merging}
UzZaman, N., H.~Llorens, and J.F. Allen (2012), \enquote{{Merging Temporal
  Annotations}.} In \emph{Proceedings of the TIME Conference}.
\bibAnnoteFile{uzzaman2012merging}

\bibitem[{Verhagen et~al.(2009)Verhagen, Gaizauskas, Schilder, Hepple,
  Moszkowicz, and Pustejovsky}]{TempEval1}
Verhagen, M., R.~Gaizauskas, F.~Schilder, M.~Hepple, J.~Moszkowicz, and
  J.~Pustejovsky (2009), \enquote{{The TempEval challenge: identifying temporal
  relations in text}.} \emph{Language Resources and Evaluation}, 43, 161--179.
\bibAnnoteFile{TempEval1}

\bibitem[{Verhagen et~al.(2010)Verhagen, Sauri, Caselli, and
  Pustejovsky}]{TempEval2}
Verhagen, M., R.~Sauri, T.~Caselli, and J.~Pustejovsky (2010),
  \enquote{{SemEval-2010 task 13: TempEval-2}.} In \emph{Proceedings of the 5th
  International Workshop on Semantic Evaluation}, 57--62, Association for
  Computational Linguistics.
\bibAnnoteFile{TempEval2}

\end{thebibliography}

\end{document}